\documentclass[conference]{IEEEtran}
\IEEEoverridecommandlockouts
\usepackage{cite}
\usepackage{amsmath,amssymb,amsfonts}
\usepackage{algorithmic}
\usepackage{graphicx}
\usepackage{textcomp}
\usepackage{xcolor}
\usepackage{lipsum}
\usepackage{dblfloatfix}
\usepackage{tabularx}

\graphicspath{{./figures/}}

\def\BibTeX{{\rm B\kern-.05em{\sc i\kern-.025em b}\kern-.08em
    T\kern-.1667em\lower.7ex\hbox{E}\kern-.125emX}}
    
\definecolor{review}{rgb}{0.0, 0.0, 0.0}

\begin{document}

\title{A Temporal Graph Neural Network for Cyber Attack Detection and Localization in Smart Grids}
\author{\IEEEauthorblockN{Seyed Hamed Haghshenas}
\IEEEauthorblockA{\textit{Dept. of Electrical Engineering} \\
\textit{University of South Florida}\\
Tampa, FL, USA \\
seyedhamedhaghshenas@usf.edu}
\and
\IEEEauthorblockN{Md Abul Hasnat}
\IEEEauthorblockA{\textit{Dept. of Electrical Engineering} \\
\textit{University of South Florida}\\
Tampa, FL, USA \\
hasnat@usf.edu}
\and
\IEEEauthorblockN{Mia Naeini}
\IEEEauthorblockA{\textit{Dept. of Electrical Engineering} \\
\textit{University of South Florida}\\
Tampa, FL, USA \\
mnaeini@usf.edu}
}

\maketitle

\begin{abstract}
This paper presents a Temporal Graph Neural Network (TGNN) framework for detection and localization of false data injection and ramp attacks on the system state in smart grids. Capturing the topological information of the system through the GNN framework along with the state measurements can improve the performance of the detection mechanism. The problem is formulated as a classification problem through a GNN with message passing mechanism to identify abnormal measurements. The residual block used in the aggregation process of message passing and the gated recurrent unit can lead to improved computational time and performance. The performance of the proposed model has been evaluated through extensive simulations of power system states and attack scenarios showing promising performance. The sensitivity of the model to intensity and location of the attacks and model's detection delay versus detection accuracy have also been evaluated.\\
\end{abstract}

\vspace{-0.5cm}
\section{Introduction}
\label{sec1}

Smart grids are being extensively equipped with sensing and monitoring devices to improve their performance and reliability. For instance, phasor measurement units (PMUs) are deployed in smart grids and designed to acquire the physical measurements of the system. The measurements will be relayed over communication networks to enhance situational awareness and the operation of the system \cite{das22}. The cyber elements of smart grids provide new opportunities for improving the operation and control of these systems; however, they also introduce vulnerabilities to cyber attacks.

The measurement data in the smart grid that are collected through the sensors and communicated over the communication networks, such as voltage, current, power injections, and the status information of breakers and switches are vulnerable to cyber attacks. These data are utilized to obtain the states of the power system, which is crucial for the proper operation and maintenance of the smart grid to ensure a seamless supply of electricity to the consumers. Cyber attacks, by disrupting the availability and integrity of the measurement data, can threaten the reliability and performance of smart grids, and their timely detection and localization are important from the grid operators' perspective.  In this work, two of the cyber attacks related to the integrity of the smart grid data, namely False Data Injection Attacks (FDIAs) \cite{liu11,musleh20,hasnat22} and a special form of FDIA, namely, Ramp attacks \cite{chu19,hasnat21,hasnat22} are considered. Both attacks aim to stealthily change data in the system measurements such that it affects functions that rely on them.

To capture the topological structure and the connectivity information of the smart grid along with the temporal measurement data, power system measurements can be modeled as time-varying graph signals \cite{ramakrishna21,hasnat22} so that they can be analyzed by graph signal processing tools which extends the concepts and tools of classical signal processing to irregular graph domain data in non-euclidean space \cite{ortega18}. Moreover, graph neural network (GNN)-based analyses can also be applied to the data modeled as time-varying power system graph signals \cite{boyaci22jan,boyaci22jun,hossain21}. In this work, a temporal GNN (TGNN) framework is presented to consider both the temporal measurements at each bus of the system and the topological connectivity (by considering the buses as the vertices/nodes, and transmission lines as the edges of the graph) to detect and locate the attacks. Specifically, the detection and localization problem is formulated as a classification problem through a GNN with a message passing mechanism to identify abnormal measurements at buses. A Gated Recurrent Unit (GRU) is implemented in the model to capture the temporal features from the time-series data on each bus in the power system. Additionally, a residual block is designed to address the vanishing gradient problem and optimized the spatio-temporal features acquired by the GNN.

The performance of the proposed model has been evaluated through the training and testing of the proposed TGNN method with FDIA and ramp attack data specially designed for time-series data from the power system\cite{hasnat22} that does not contain abrupt changes of values at the onset making them challenging to detect. The results show a promising performance with high accuracy in detecting and locating both the FDIA and ramp attacks. The sensitivity of the model to the intensity of the attacks and the topological location of the attacks have also been evaluated suggesting that detecting attacks at certain locations are more challenging than the other locations.

\section{Related WorK}
\label{sec2}
Smart grid's security have gained considerable attention from researchers and practitioners in the past few decades. Both model-based and data-driven techniques for the detection and localization of cyber attacks are proposed and studied in the literature \cite{musleh20}. While model-based approaches \cite{musleh20,liu11} require the knowledge of the system and information on its structure and dynamics, the data-driven approaches utilize the large volume of available measurement data to detect cyber attacks. The majority of the data-driven detecting and locating techniques are signal processing-based \cite{chu19}, machine learning-based \cite{zhang21}, and their combinations \cite{hasnat21}. However, one of the limitations of these techniques is that they do not explicitly capture and utilize the underlying structure of the system in their analyses.
As smart grids have a graph structure that reflects the connectivity of their components and can affect the dynamics and interactions of the components, the data from these systems naturally have embedded structures. Therefore, by modeling the structure of the power grid as a graph and the data associated with it as graph signals, one can capture the information related to the connectivity and interdependence among the components of the grid. The power system measurement data modeled as graph signals can be utilized in two ways for detecting cyber attacks: 1) by applying graph signal processing tools to the power system graph signal \cite{drayer20,ramakrishna21,hasnat22}, or 2) using graph neural networks \cite{boyaci22jan,boyaci22jun} to the graph-structured data. A common GSP-based approach is to utilize the graph Fourier transform (GFT) to analyze the presence of a high-graph frequency component as the indicator of falsified measurements \cite{drayer20,ramakrishna21,hasnat22}. However, the \textit{local smoothness second time-derivative (LSSTD)}-based method \cite{hasnat22}, by capturing both the temporal and vertex-to-vertex evolution of the graph signal is effective for detecting and locating sophisticated designed cyber attack with no sharp changes of signal values at the onset of the attack. The main limitation of the GSP-based method is the requirement of decision thresholds to be applied to the graph-spectral measures, which can be challenging to fix empirically. On the other hand, GNN-based methods automatically adjust the model parameters to utilize graph-structured data along with the topology. Detection and localization of cyber attacks in smart grids can be cast into a node classification problem within the GNN framework. For instance, in \cite{boyaci22jan,boyaci22jun}, GCN frameworks have been used for FDIA detection to classify normal buses in the system from the ones with malicious data. 
The current paper presents a spatio-temporal-GNN framework for cyber attack detection and location identification in smart grids, which also models the task as a node classification problem. 
Compared to the work presented in \cite{boyaci22jan} based on GNN, the model presented in the current paper uses a combination of message passing process and GRUs in residual blocks in the places of graph convolutional layers and auto-regressive moving average (ARMA) filter, respectively, which can improve the computational time and performance.

\section{METHODOLOGY}
\label{sec3}
\subsection{Power System Model}
In this paper, the physical topology of the power system is represented as a graph $G:=\{\mathcal{V},\mathcal{E}\}$ with adjacency matrix, $\mathbf{A}=[a_{ij}]$, where $\mathcal{V}$ is the set of vertices of the graph representing the buses, and $\mathcal{E}=\{e_{ij}: (i,j) \in \mathcal{V} \times \mathcal{V}, a_{ij}=1\}$ is the set of all edges representing the transmission lines of the power grid. Various attributes can be associated with each bus $n\in \mathcal{V}$, for instance, the real and reactive power injections, bus voltage magnitude and angle, injected bus current magnitude and angle, frequency, etc. This work considers the real power injection at each bus as a time-varying graph signal, $x(n,t)$. 
Specifically, the signal value of $x(n,t)$ represents the injected real power at bus $n$. It is also assumed that the measurements are available at all the buses of the system (for instance through PMUs or state estimation mechanisms).
\subsection{Cyber Attack Model}
Cyber attacks on the power system measurements (e.g. SCADA or PMUs) have been modeled in different ways depending on the diversity of the scenarios, research problems, and perspectives of the research. In this work, the cyber attacks are modeled on the time-varying real power injection in each bus. The cyber attacks considered in this work are designed in such a way that they do not contain abrupt changes in signal values at the onset of the attack making them challenging to detect. For modeling the cyber attacks, i.e., FDIA and ramp attack, we endorse the generalized approach presented in \cite{hasnat22}. Specifically, let $\mathcal{V_A} \subset \mathcal{V}$ denotes the set of all buses under cyber attack at time interval $[t_{start} , t_{end}]$. The cyber attacks on the bus real power injection time series in the generalized form can be expressed by the following equation:
\begin{equation} \label{gen.cmodel}
x(n_{A}, t)=c(t),\ \text{for}\ \ t_{start}\leq t \leq t_{end},\ \ \text{and}\ \ n_A \in \mathcal{V}_{A}
\end{equation}

The cyber attacks considered in this paper can be modeled as special cases of equation (\ref{gen.cmodel}) as discussed next. 

\subsubsection{False Data Injection Attack}
There are various work on modeling FDIA in smart grids and the goal of such models are to characterize FDIA that can bypass the bad data detectors and affect the state estimation function \cite{ramakrishna21}. Considering the state estimation framework in power systems with $\bf{z}=\bf{h}(\bf{y})$, where $\bf{z}$ and $\bf{y}$ are the measurements and the states of the power system and $\bf{h}$ is a non-linear function, which relates measurements and states, the goal of the FDIA is generally to inject bad data to the set of measurements to compromise the performance of the estimator. If the residue of state estimation $r={||\bf{z}-\bf{h}(\bf{\hat{y}})||}_2$ error is larger than a defined threshold, where $\bf{\hat{y}}$ is the estimated states then a bad data can be detected. If the attack is designed in a way that passes the aforementioned test then the bad data detector cannot detect the attack. 
Following the work in \cite{hasnat22}, in this work, FDIA with no sharp change at the onset of the attack is modeled by considering $c(t)=x(n_A,t)+(-1)^b x'$, in equation (\ref{gen.cmodel}) where $b \in \{0,1\}$, $|x'|$ is considered to be a very small value, which does not result in sharp changes in the onset of the attack and bypasses the bad data detector as discussed earlier.

\begin{figure*}[!]
    \centering
    \vspace{-0.3cm}
    \includegraphics[width=19cm, height=3cm]{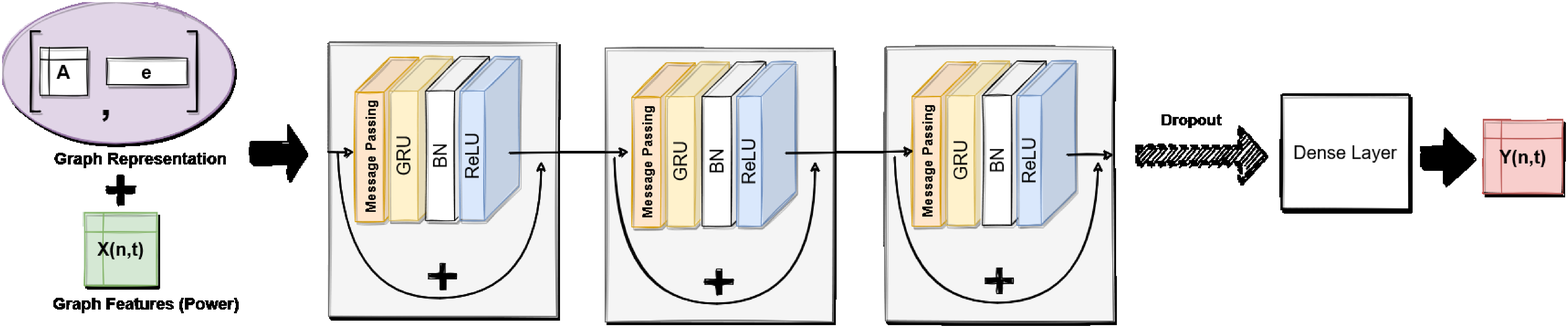}
    \caption{Architecture of the proposed TGNN for real-time cyber attack detection and localization.}
    \label{fig0:architecture}
    \vspace{-0.5cm}
\end{figure*}

\subsubsection{Ramp Attack}
{\color{review} {Ramp attack is a special form of FDIA.}}
In ramp attack, to ensure the smooth changes of values at the attack onset, falsified measurements are inserted gradually into the bus(es) under cyber attack, which makes them challenging to detect. Ramp attack can be modeled as  $c(t)=x(n_A,t_{start}) + m \times (t-t_{start})+q(n_A,t)$, where $m$ is the slope and $q(n_A,t)$ is the additive white Gaussian noise associated with the measurement devices at the bus $n_A$.

\color{black}
\subsection{Model Architecture}
In this section, the TGNN model used in this paper is introduced. {\color{review}{
Given the adjacency matrix $A:=\{0,1\}\in R^{\{N\times{N}\}}$, and feature matrix $X$, a GNN layer can be expressed as $H^{l+1}=\sigma({\Tilde{D}^{-1/2}}\Tilde{A}{\Tilde{D}^{-1/2}}H^{l}W_{l})$. Here, $\Tilde{A}:=A+I_{N}$ (where $I_{N}$ is the identity matrix of size $N$) and $\Tilde{D}$ is the degree matrix. $\sigma(.)$ is the sigmoid activation function, $I$ is the layer number, $W_{l}$ holds the weights of layer $l$ and $H^{l}$ is the output of layer $l$. To create a multi-layer GNN, multiple of such layers can adopted.}}
Fig.\ref{fig0:architecture} demonstrates the architecture of the proposed TGNN model in this paper. This model is following the message passing method of general GNNs \cite{wu21}. The proposed architecture consists of three sequential layers aiming to extract and optimize the spatio-temporal features of the data following by a dropout and a dense layer. Each of the three layers contains; 1) A message passing block to acquire the structural and topological features of the data, 2) A GRU block to obtain the temporal features of the data, 3) A residual block in which Batch Normalization and a Rectified Linear Unit (ReLu) activation function is applied to the data and 4) A skip connection that feeds the output of one layer as the input to the next layer.

The message passing process can be describe as following. At node-level, the embedding of node $i$ will be updated based on the aggregated feature of the node itself along with its one-hop neighbors as $f(x_i)=\phi(x_i, \oplus_{j\in N_{i}}\psi(x_i,x_j))$, where $N_i$ is the set of one-hop distance neighbors of node $i$, $\oplus$ is the permutation invariant aggregation function and $\phi$ and $\psi$ are learnable functions to be characterized using neural network. The three sequential GNN layers allow message passing up to three-hop distance neighbors. Afterwards, the GRUs are implemented in our architecture. GRUs are improved version of standard recurrent neural network. They aim to solve the vanishing gradient problem \cite{cho14}. Additionally, due to their update gate and reset gate feature, they can keep information from long ago, without washing it through time.\\
Every GRU has four gates. 1) \textit{Update gate($z_t$)}: for time step $t$ using the $z_t=\sigma(W^{(z)}{x_t} + U^{(z)}h_{t-1})$. When $x_t$ is fed into the unit, it is multiplied by its weight $W^{(z)}$. The same applies to $h_{t-1}$ that contains the information of the previous $t-1$ units and is multiplied by its own weight of $U^{(z)}$. Then a Sigmoid activation function is applied to squash the results between $0$ and $1$. The update gate determines how much of the past information(from previous time steps) needs to be passed along to the next unit. 2) \textit{Reset gate($r_t$)}: this gate decides how much of the past information to forget. It follows the formula similar to the previous one, $r_t=\sigma(W^{(r)}{x_t} + U^{(r)}h_{t-1})$ with a difference in the weights and the gate's usage in the next gates. 3) \textit{Current memory content ($h'_t$)}: it will use the output of the reset gate to store the relevant information from the past. The formula is $h'_t=\tanh{(W{x_t} + r_{t}\otimes{Uh_{t-1}})}$, where the input $x_t$ and information from the previous unit are multiplied by their weights, $W$ and $U$, respectively, and the reset gate is applied to the previous unit information by Hadamard (element-wise) product ($\otimes{}$). This helps the unit to determine what to remove from the previous time steps. At the end, these two calculated results are summed up and a non-linear activation function of $\tanh$ is applied to them. 4) \textit{Final memory at current time step ($h_t$)}: $h_t$ is the vector that holds information for the current unit and passes it down to the network. To calculate the current time step memory, the update gate is needed since $h_t=z_t\otimes{h_{t-1}} + (1-z_{t})\otimes{h'_{t}}$. Based on the formula, the unit can learn to set the $z_t$ close to $1$ and keep a majority of the previous information. $z_t$ being close to $1$ causes the ($1-z_t$) being close to $0$ that ignores big portion of the current information. The sequence of mentioned four gates in GRU allows it to store and filter the information using the update and reset gates. Therefore, the GRU is able to eliminate the vanishing gradient problem.

Once the spatial and temporal features extracted, residual blocks and skip connections are applied. They aim to address the degradation problem and feature reusability. The results are then fed to a dropout and a dense layer which produces softmax values to the output layer. The model produces either $0$ or $1$ labels for the nodes of the system at each time instance specifying if the measurement is normal or abnormal, respectively. Therefore, the TGNN output specifies the beginning and ending time of the detected abnormal data as well.


\subsection{Model Parameters}
In this work, Stochastic Gradient Descent (SGD) optimizer is used in our TGNN model. The learning rate of $0.001$, the exponential decay rate of $e^{(-6)}$, momentum of $0.9$ that accelerates gradient descent in the relevant direction and dampens oscillations, and Nesterov momentum are considered. {\color{review}{Nesterov momentum calculates the decaying moving average of the gradients of projected positions in the search space as a substitute to the actual positions \cite{nesterov1, nesterov2}.}} The size of hidden layers in the dense output layer is set to $64$ and the dropout rate is set as $0.3$. Threshold value of $0.5$ is used in the output layer for deciding the binary labels. 
In this work, simialar to {\color{review}{\cite{boyaci22jun, boyaci22jan}}}, the real power injections at the buses are considered as state representors of the buses or features associated with each bus. While other measurements including voltage magnitude $V_i$ and voltage angle ${\theta}_i$ can also be considered and have been considered in the literature, our experiments have shown strong correlation among these features. {\color{review}{The model takes input batch size of $256$ representing the number of samples processed before the model is updated.}} \color{black} The binary cross entropy is used as the loss function. The network monitors the binary accuracy, which determines the ratio of correct predictions of labels.
\vspace{-0.1cm}
\section{Results}
\label{sec4}
\subsection{Data Generation and Data Processing}
In this study, the time-series power system measurements are obtained by running power flow in MATPOWER 7.0 \cite{matpower} on IEEE 118 bus system \cite{118bus} by varying the load demands in time according to the load pattern obtained from the New York Independent System Operator (NYISO) \cite{nyiso} as described in \cite{hasnat22}. {\color{review}{The system state \textit{$x$} (\textit{$V_i$} and \textit{$\theta_i$} at each bus) is estimated using the PSSE module \cite{boyaci22jan, boyaci22jun}. PSSE solves the optimization problem in $\hat{x}=\min_{x}(z-h(x))^{T}R^{-1}(z-h(x))$, as a weighted least squares estimation using complex power measurements $z$ collected by PMUs. In the optimization equation, $R$ represents measurements' error covariance matrix and $z$ includes $P_{i}, Q_{i}, P_{ij}$ and $Q_{ij}$.}} We consider the bus real power injections as the measurement data with $52,500$ time instances for training and $17,500$ time instances for testing. For half of the instances, cyber attacks (FDIA, and ramp) are launched according to the model described in Section III.B. $15\%$ of training data have been used for cross-validation. {\color{review}{The measurement data sampling frequency is $30$Hz.}} As mentioned earlier, the maximum number of epochs for training considered is set to be $100$ with early stopping criteria, where $75$ epochs are tolerated without any enhancement in the accuracy and loss of validation set. All the implementations are executed in Python$3.6$ using Sklearn, Tensorflow, Keras, and Networkx on an Intel Core i7-7700 CPU $3.60$GHz. 

\subsection{Detection Performance of TGNN Framework}
\begin{figure}[h]
    \centering
    \vspace{-0.1cm}
    \includegraphics[height=1.5in]{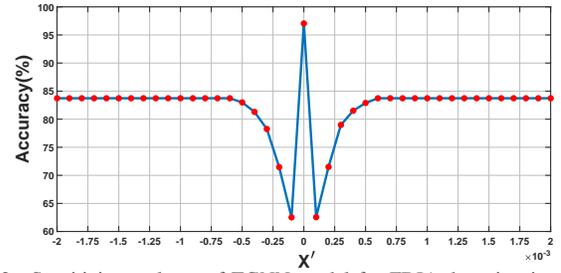}
    \vspace{-0.5cm}
    \caption{Sensitivity analyses of TGNN model for FDIA detection in terms of accuracy for various intensities of attack, $|x'|$.}
    \label{fig2:sens_fdia}
    \vspace{-0.3cm}
\end{figure}

The performance of the proposed TGNN Framework is evaluated in terms of detection accuracy and detection delay. The latter metric is important for the suitability of implementing the technique in real-time scenarios. The TGNN model detects FDIA and ramp attacks with an overall accuracy of $99.50\%$ and $88.85\%$, respectively.
{\color{review}{A sensitivity analysis has been conducted to evaluate the role of the intensity of the FDIA attack on the performance of the TGNN model. Specifically, the intensity of the FDIA has been varied by considering various $|x'|$ values in the range of [$-0.002$, $0.002$] per unit as discussed in Section~III. The results of this sensitivity analysis in terms of the acquired accuracy are presented in Fig.~\ref{fig2:sens_fdia}. When $|x'|=0$, which represents no attack, the model performance is high in labeling no attacks in the system. When $|x'|=0.0001$ then the TGNN model accuracy drops to $62.51\%$, which is the point where the model fails to detect the attack.}}


The results indicate that TGNN can detect and locate the FDIAs exactly at the instances in which that attacks are introduced. However, for the ramp attacks, the detection delays are in the range of $0$ to $20$ time instances, where most of the attacks are detected within $9$ time instances while for only a few cases the delay is large (around $20$ time instance). For this reason, the median detection delay is chosen as the metric for the real-time applicability of the TGNN model.

\begin{figure}[h]
    \vspace{-0.3cm}
    \includegraphics[width=9cm ,height=5cm]{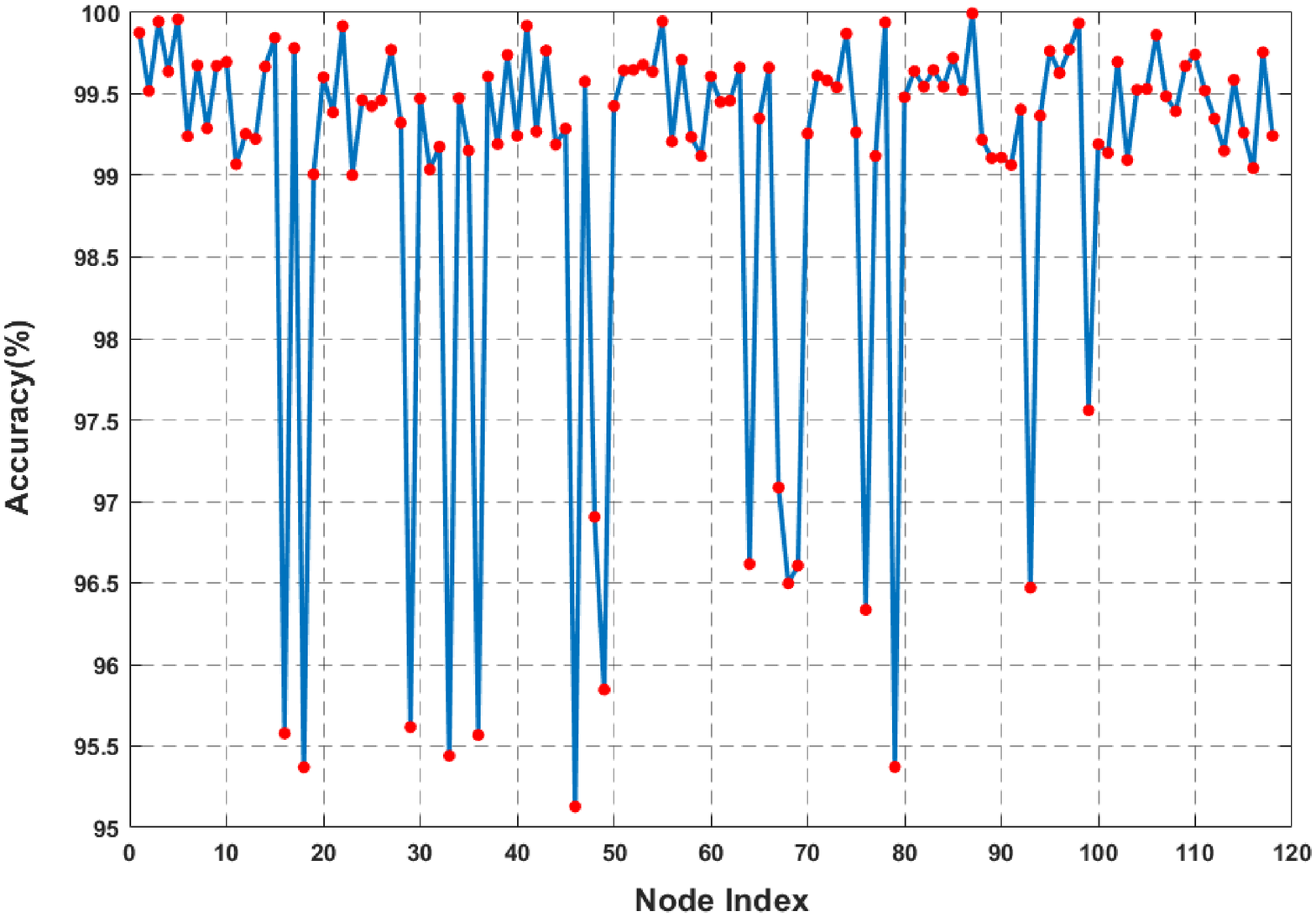}
    \vspace{-0.7cm}
    \caption{Sensitivity analyses of TGNN model for FDIA detection in terms of accuracy for various locations (i.e., bus index) of the attack.}
    \label{fig3:nba}
    \vspace{-0.5cm}
\end{figure}

\begin{figure}[h]
    \centering
    \includegraphics[height=1.5in]{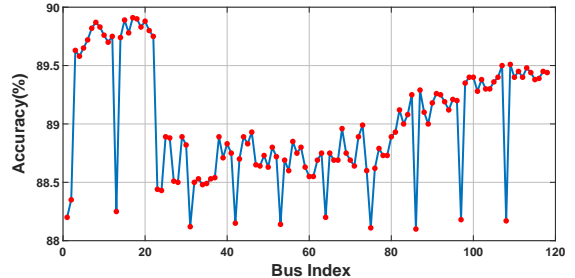}
    \vspace{-0.2cm}
    \caption{Sensitivity analyses of TGNN model for ramp attack detection in terms of accuracy for various locations (i.e., bus index) of the attack.}
    \label{fig4:ramp_nba}
\end{figure}

In order to evaluate the impact of FDIA in different buses, $1000$ FDIA has been simulated and considered separately on each node with $x'=0.02$ on a one-day, i.e., 2,500 instance dataset. The accuracy of the FDIA detection in each of the buses are shown in Fig.~\ref{fig3:nba}. It can be observed that some of the buses of the system are more sensitive to attacks than others by reflecting the lower performance of detection. 
This suggests that the proposed model can be sensitive to the location of the FDIA. Similar analysis have been performed for the ramp attack as illustrated in Fig.~\ref{fig4:ramp_nba}. {\color{review}{The accuracy in some of the buses drops to $88.1\%$ while in others it is up to $89.91\%$.}}

\begin{figure}[h]
    \centering
    \vspace{-0.1cm}
    \includegraphics[height=1.5in]{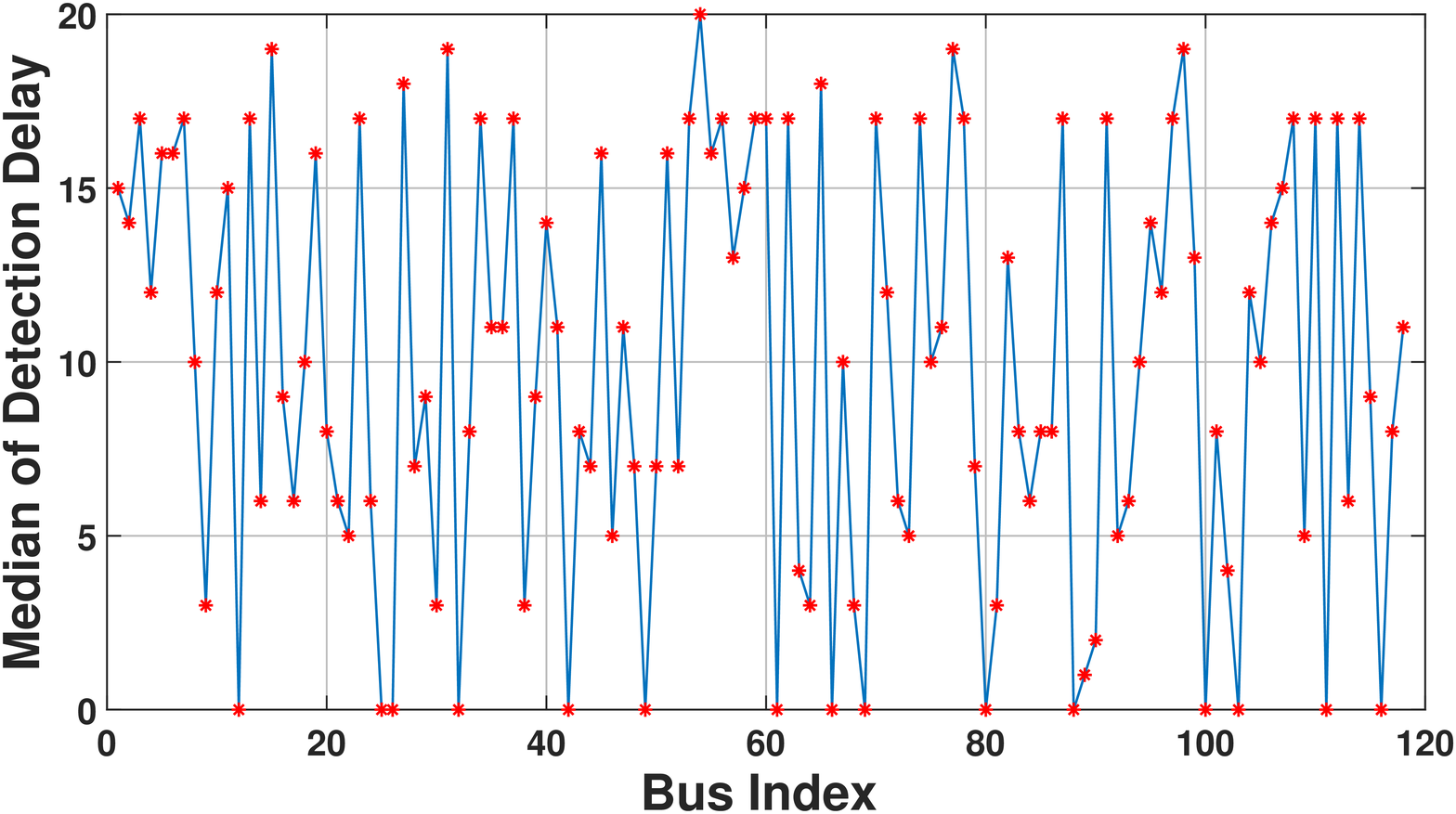}
    \vspace{-0.5cm}
    \caption{Sensitivity analyses of TGNN model for ramp attack detection in terms of median of detection delay for various locations.}
    \label{fig5:ramp_dd}
    \vspace{-0.4cm}
\end{figure}

\vspace{-0.4cm}
\subsection{Comparison with baseline GNN model}
\begin{figure}[h]
    \centering
    \vspace{-0.1cm}
    \includegraphics[height=1.5in]{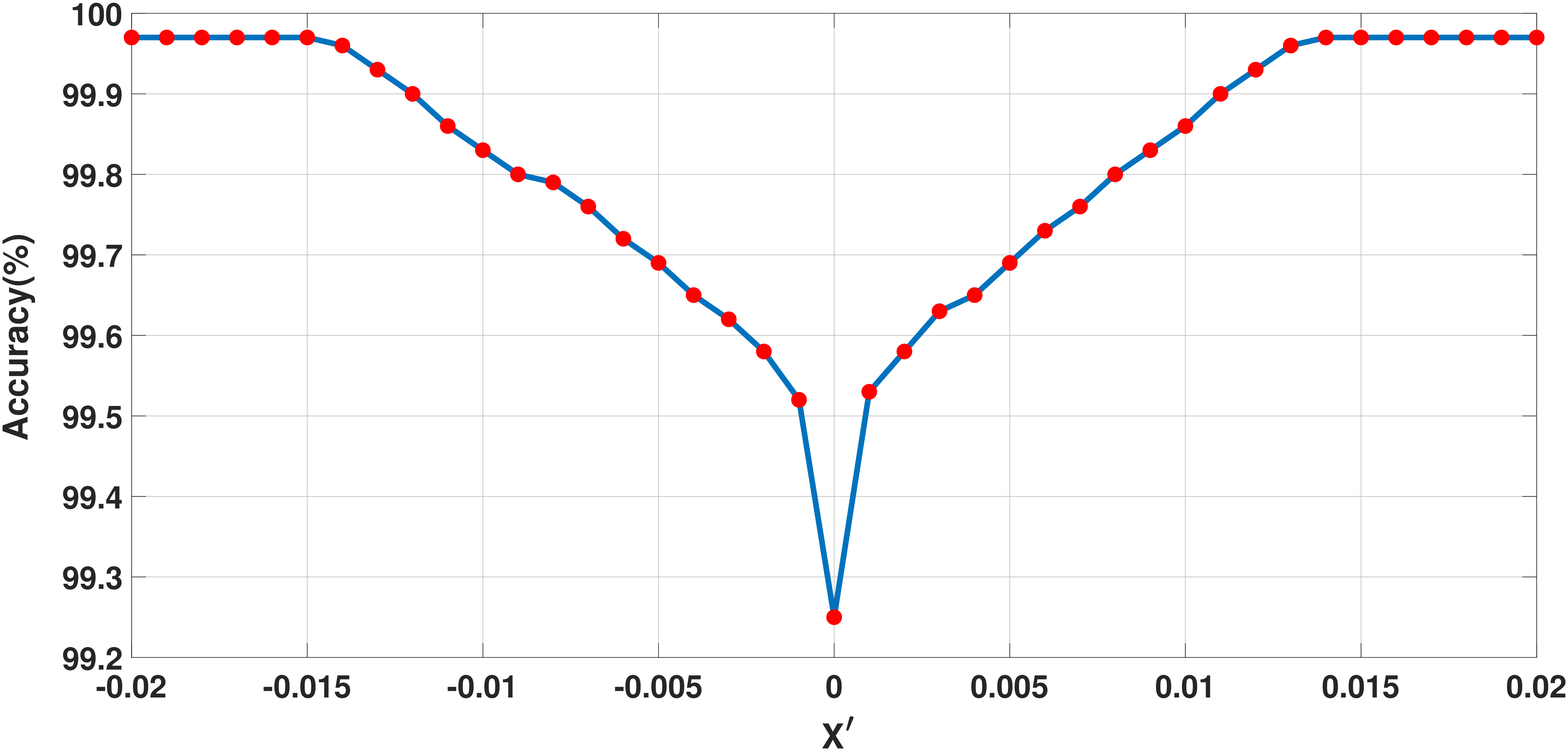}
    \vspace{-0.3cm}
    \caption{Sensitivity analyses of GNN model for FDIA detection in terms of accuracy for various intensities of attack, $|x'|$.}
    \label{fig6:sens_gnn}
    \vspace{-0.3cm}
\end{figure}

We have also developed a baseline GNN model, which does not consider the temporal information and therefore, is mainly suitable for detecting and locating attacks from snap-shot data. It comprises a message passing block similar to the message passing module introduced in the TGNN model and a Feed-Forward Network (FFN) block that prepares and updates the spatial features at each bus of the power system. Applying this baseline GNN model to the FDIA data, {\color{review}{high accuracy of $99.97\%$ can be achieved. A similar sensitivity analysis for the intensity of the attack $|x'|$ for this model has been illustrated in Fig.~\ref{fig6:sens_gnn} in the range of [$-0.02$, $0.02$]. This baseline GNN model fails to detect FDIAs at $|x'|=0.001$, which is $10$ times larger than the failure point of the proposed TGNN model}}. The improved performance of TGNN is due to exploiting the temporal information and the residual blocks that allow dynamic tuning of the parameters during training. Finally, the F1 score and the false alarm rate for the proposed TGNN model is $100\%$ and $0.00\%$ for FDIA and $99.98\%$ and $0.02\%$ for ramp attacks, respectively. The F1 score and FA of GNN model for FDIAs is $100\%$ and $0.00\%$, subsequently. However, TGNN can facilitate the detection and locating of ramp attacks that cannot be detected by the baseline GNN model.


\vspace{-0.2cm}
\section{Conclusion}
\label{sec5}
In this work, a TGNN framework has been proposed that uses both the temporal and topological information from the system to detect and localize the cyber attacks. Adopting the message passing method, GRU, and residual blocks in the model, the proposed TGNN is capable of producing and processing messages and updating the node embeddings by aggregating and summing node representations in each node's neighborhood. The model can detect and localize FDIAs and ramp attacks with high accuracy, which was verified experimentally using simulation data on the IEEE 118 bus system. Moreover, sensitivity analyses of the model based on different magnitudes of attack and the location of the attack were performed and compared to our baseline GNN model that is developed to only take the topological structures of the power system along each snap-shot of the states. 

\section{Acknowledgement}
\vspace{-0.1cm}
This material is based upon work supported by the National Science Foundation under Grant No.~2118510.

\end{document}